\documentclass[letterpaper, 10 pt, conference]{ieeeconf}  

\IEEEoverridecommandlockouts                              

\usepackage{amssymb}
\usepackage{graphicx}
\usepackage{float}
\usepackage{amsmath}
\usepackage{booktabs}
\usepackage{placeins}

\title{\LARGE \bf
Efficient Transfer Learning of Robot Dynamic Models Using Morphological Similarity
}

\author{Pavlo Kupyn$^{1, 2}$, Yuya Hamamatsu$^{1}$, Roza Gkliva$^{1}$, Asko Ristolainen$^{1}$, Maarja Kruusmaa$^{1}$
\thanks{$^{1}$The authors are with the Department of Computer Systems, Tallinn University of Technology, Tallinn, Estonia
        {\tt\small (Pavlo.Kupyn, Yuya.Hamamatsu, Roza.Gkliva, Asko.Ristolainen,
        Maarja.Kruusmaa)@taltech.ee}, $^{2}$ University of Bonn, Institute of Computer Science, Bonn, Germany}
}

\begin{document}

\maketitle
\begingroup
\renewcommand{\thefootnote}{}
\footnotetext{\scriptsize \copyright~2026 IEEE. Personal use of this material is
permitted. Permission from IEEE must be obtained for all other uses. Accepted at
CoDIT 2026.}
\endgroup

\thispagestyle{empty}
\pagestyle{empty}

\begin{abstract}

This study proposes a neural network–based transfer learning framework for modeling the dynamics of soft, fin-actuated underwater robots. We focus on morphologically similar robots that differ in scale and hydrodynamic properties. A model trained on data from a larger robot (source domain) is adapted to a smaller one (target domain) with limited labeled data. To enable label-efficient transfer, we develop an autoencoder-based domain adaptation approach that learns a shared latent representation aligning the dynamics of both robots. Experiments on two real underwater robots show that the proposed method enables accurate state estimation of the body-frame velocities on a target platform without labeled data, highlighting its potential for efficient cross-robot dynamics transfer among morphologically similar platforms.

\end{abstract}


\section{Introduction}

The morphologies of the underwater robots vary by application \cite{biomimetics8030318}. The limitations of conventional propeller-driven systems \cite{Zhang_2024_opt} motivate bio-inspired robots to use soft fin-based actuators that offer quieter and efficient propulsion. However, accurately modeling the dynamics of such soft actuated systems for state estimation and control remains challenging due to complex nonlinear fluid–structure interactions \cite{zheng2022three}. Due to this difficulty, most underwater vehicle systems currently rely on state estimation via expensive sensor feedback \cite{wu2024review}.
To address these challenges, Neural Network (NN)-based modeling and state estimation have shown promise in capturing these complexities by predicting robot motion directly from sensor and control input. However, they often require large labeled datasets for supervised training \cite{singh2025deepvldynamicsinertialmeasurementsbased}. Acquiring such data for underwater platforms is costly and difficult due to the requirements of special equipment to obtain ground truth labels.

To address these problems, transfer learning can leverage knowledge from one robot to improve learning on another \cite{jaquier2024transferlearningroboticsupcoming}. In particular, this work explores dynamics transfer between two morphologically similar, fin-actuated underwater robots. Although both share the same actuation principles and kinematics, differences in geometrical scale introduce a domain gap. 
\begin{figure*}[t]
    \centering
    \includegraphics[width=\linewidth]{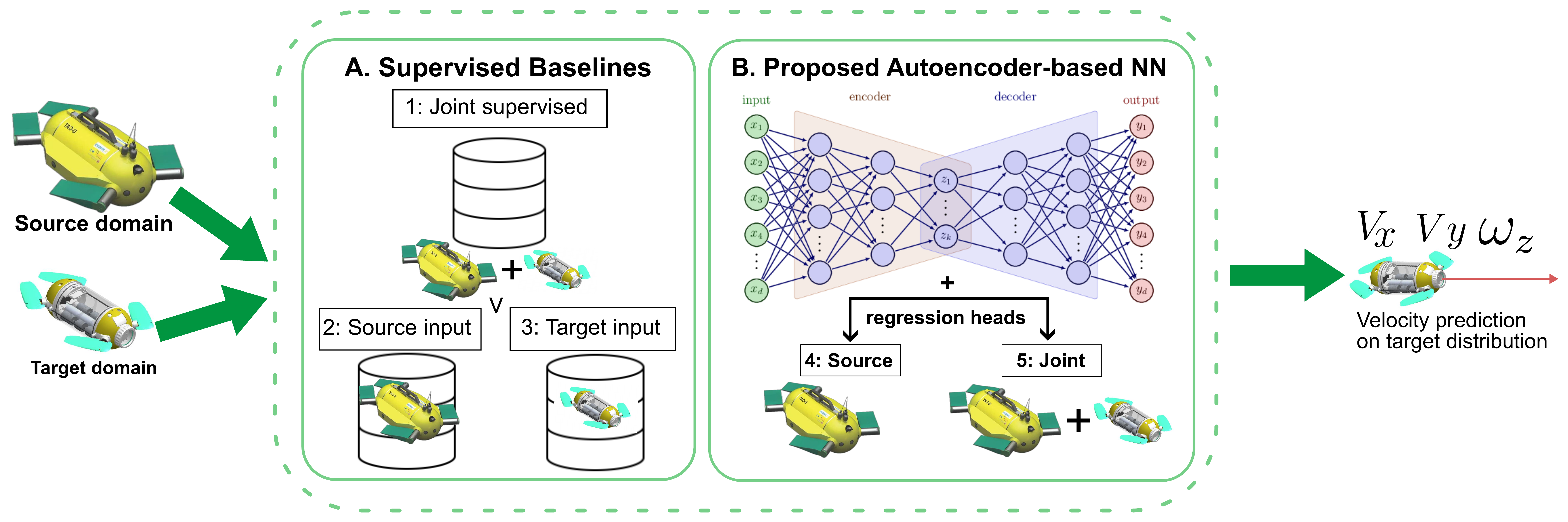}

    \caption{Overview of compared approaches. (A) Supervised baselines: (1) joint, (2) source-only, (3) target-only. (B) Proposed autoencoder with (4) source-only and (5) joint regression heads.}
    \label{fig:overview}
\end{figure*}
To bridge this gap, we systematically evaluate several dynamic modeling transfer strategies. First, we evaluated supervised baseline approaches, including neural networks trained independently on each robot’s dataset, as well as a joint model trained on combined dataset. Then, we propose a self-supervised autoencoder framework that learns a shared latent representation aligning both robot domains. An overview of the compared approaches is shown in Fig.~\ref{fig:overview}. By encoding the joint input feature space, the model captures each robot's dynamics and enables an accurate prediction for the smaller robot even without labeled target data. Accurate dynamic forward modeling serves as a foundation for model-based control. Improving the transferability of learned dynamics directly contributes to enhancing control in soft robot control applications. We validate the proposed framework on two morphologically similar fin-actuated underwater robots:  \textbf{U-CAT}, referred to as the source domain, and \textbf{Micro-CAT}, serving as the target domain. 

\section{Related Works}

Soft-actuated underwater robots are difficult to model due to nonlinear fluid–structure interactions \cite{muhammad2014non, singh2019dynamic}. While analytical models fail to capture these effects \cite{GEORGIADES200939}, neural network–based methods offer better accuracy and flexibility \cite{lee2023data, 9652036}. A surrogate model has been proposed to capture the dynamics of fin actuators \cite{hamamatsu2025underwater}, but it is limited to the single-fin case. Such data-driven models still rely on large labeled datasets, which are challenging to collect for motion-constrained underwater robots. To reduce data dependency, transfer learning has been applied to reuse knowledge across robotic platforms \cite{jaquier2024transferlearningroboticsupcoming}. For example, layer policy transfer for deep reinforcement learning control has shown to improve learning efficiency in underwater robots \cite{Hamamatsu_2025}, but it still relies on extensive labeled data. In parallel, a self-supervised domain adaptation approach has been used to bridge domain gaps, for instance, by pre-training with synthetic underwater data to enhance visual models \cite{UDP_transfer_learning2024}. However, these efforts still primarily address perception tasks.
Recent work in robotic manipulation has explored cross-embodiment transfer using aligned latent representations \cite{dastider2025crossembodimentroboticmanipulationsynthesis}. The approach proposed in this work aligns knowledge from two different domains into a common latent representation, which later can be used to guide the learner domain for the generation of robotic manipulation trajectories. This approach highlights the potential of representation alignment for knowledge transfer across different robot embodiments.

Unlike prior works that focus on visual or control policy transfer, our study targets latent dynamic model alignment between morphologically similar soft underwater robots.

\section{Supervised Baseline Models}\label{sec:baselines}

This section outlines the supervised baseline methods used to model the motion dynamics of the target platform. In total, three baseline models are evaluated. All baselines use the same neural network architecture consisting of a 1D convolutional layer, a max-pooling layer, and a fully connected output layer. The models are trained using the Adam optimizer for 50 epochs under identical conditions to the autoencoder to ensure a fair comparison. The baselines consist of standard neural networks trained on sequences of sensor and actuation features to predict robot motion. In particular, input features consist of: pressure values from pressure sensor, IMU orientation, IMU angular velocity, IMU linear acceleration (gravity aligned), fin actuation parameters as described in Section~\ref{robots_specifications}.
Each baseline is trained in a supervised regression setting using time-series inputs to predict target motion variables. The loss function is the mean squared error (MSE) between the predicted and ground-truth motion values.
Three data usage settings are considered, as illustrated in Fig.~\ref{fig:overview}: 
the \textbf{target-only model}, trained on labeled data from the target robot; 
the \textbf{source-only model}, trained on source-domain data and evaluated directly on the target domain; 
and the \textbf{joint supervised model}, trained on the combined datasets of both robots. 
These settings are used as reference points for comparison with the proposed method.

\section{Method}\label{sec: method}
This section describes the proposed autoencoder model for velocity estimation based on cross-platform transfer of underwater robot motion dynamics. The model is trained in a self-supervised manner using a shared feature space constructed from both robot domains. 
It encodes sequences of sensor and controls into a compact latent space that captures common dynamic behaviors between platforms. The input features are the same as for the baseline models. During training, the model jointly optimizes the reconstruction, dynamics, and domain-alignment objectives to ensure accurate feature reconstruction and overlapping latent distributions between domains. The network is trained using Adam optimizer with a learning rate of 0.001, a batch size of 256, and 50 epochs. Although trained on both robots, it is evaluated only on the smaller target platform to test its ability to generalize and transfer dynamic knowledge from the larger robot. Further details of the architecture and training procedure are provided in the following subsections.
\subsection{Architecture}

\begin{figure}[t]
    \centering
    \includegraphics[width=\linewidth]{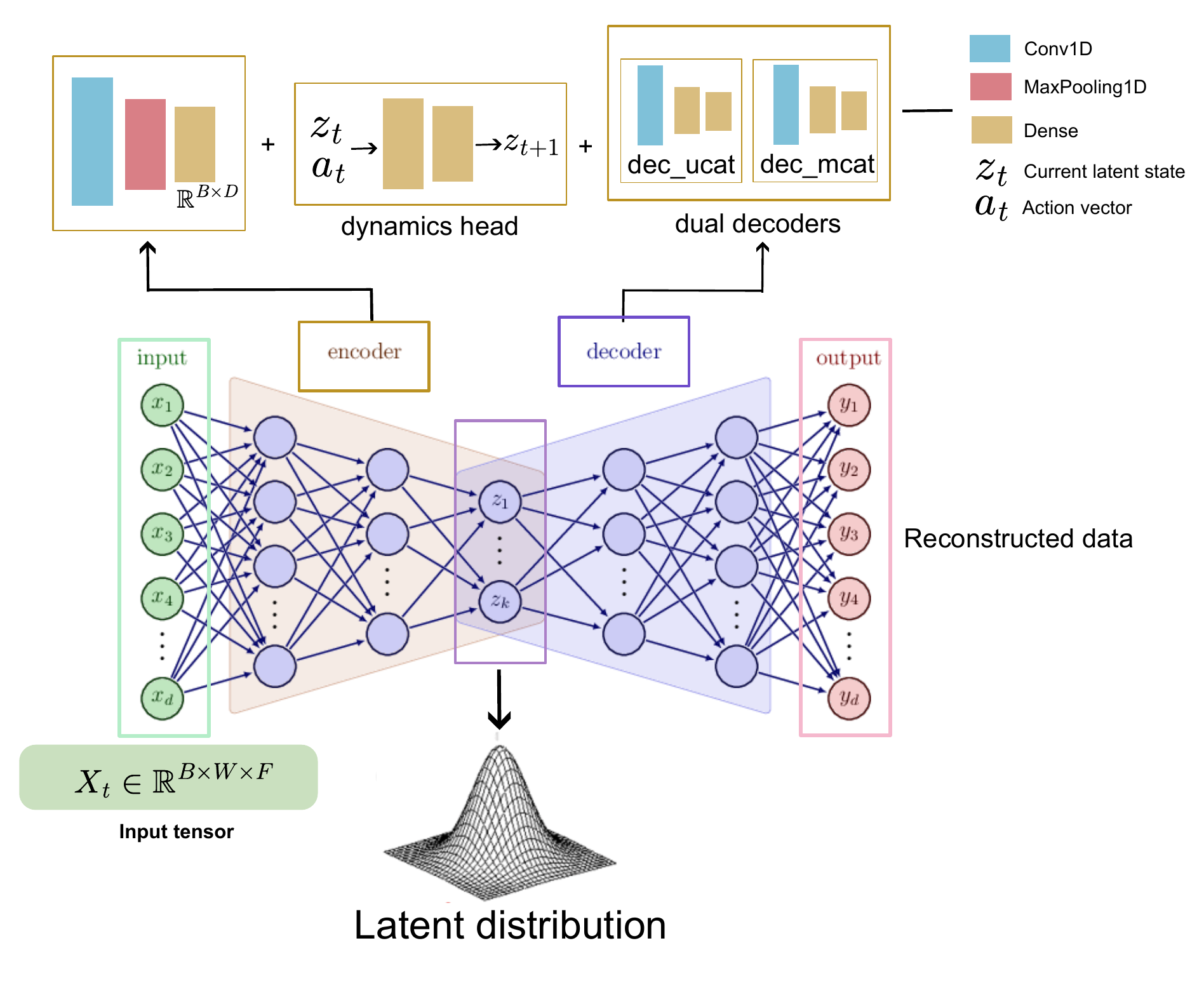}
    \caption{Detailed architecture of the proposed autoencoder}
    \label{fig:ae_architecture}
\end{figure}

\begin{figure*}[t]
    
    \includegraphics[width=\linewidth]{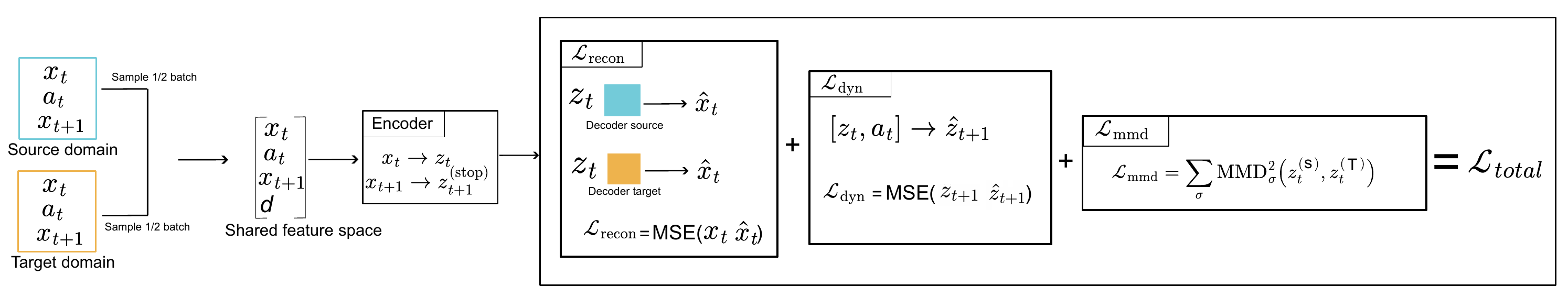}
    \caption{Autoencoder training pipeline overview, where:\\
    \hspace{0.1cm} $x_t$ – input feature tensor;
    \hspace{0.1cm} $d$ – domain identifier (0 = source, 1 = target);
    \hspace{0.1cm} $z_t$ – current latent state;
    \hspace{0.1cm} $a_t$ – action vector;
    \hspace{0.1cm} $\hat{x}_t$ – reconstructed input;
    \hspace{0.1cm} $\hat{z}_{t+1}$ – predicted next latent state;
    \hspace{0.1cm} $\sigma$ – Gaussian kernel bandwidth parameter. 
}
    \label{fig:training_pipeline}
\end{figure*}


The autoencoder architecture is illustrated in Fig.~\ref{fig:ae_architecture}. It consists of three main components: an encoder, a dynamics head, and two domain-specific decoders. The encoder extracts compact latent representations from input feature space. It consists of a 1D CNN layer that captures short-term temporal patterns, followed by pooling and dense layers that compress the input into a lower-dimensional latent vector encoding robot-specific dynamic properties. Two decoders are used to reconstruct the original time-series sequence—one per domain. Each sample is routed through the corresponding decoder based on its domain label, allowing the shared encoder to maintain reconstruction quality while supporting domain alignment. Each decoder contains two fully connected layers for reconstruction from latent representation.  The dynamics head predicts the next latent state from the current latent state and control inputs.  

\subsection{Training pipeline}
The training pipeline is illustrated in Fig.~\ref{fig:training_pipeline}.  
The data preparation stage involves constructing a shared feature space that combines sensor readings and actuation-related kinematic parameters from both domains as described in Section~\ref{robots_specifications}. For each training sample, the corresponding control input and the feature sequence of the next timestep are also generated. This results in a joint dataset consisting of feature sequences, control inputs, next-state sequences, and a domain label for source and target samples.  

The encoder then processes the preprocessed time-series data (structured as 3D tensors after windowing) and compresses them into a compact latent representation. To jointly learn feature reconstruction, temporal dynamics, and cross-domain alignment, the model is optimized using a combined loss function:
\begin{equation}
    \mathcal{L}_{total} = \mathcal{L}_{recon} + \lambda_{dyn}\mathcal{L}_{dyn} + \lambda_{mmd}\mathcal{L}_{mmd}
\end{equation}
where $\mathcal{L}_{recon}$ denotes the reconstruction loss, $\mathcal{L}_{dyn}$ the dynamics prediction loss, and $\mathcal{L}_{mmd}$ the domain alignment loss. The weighting factors $\lambda_{dyn}$ and $\lambda_{mmd}$ were both set to 0.5 to balance reconstruction, dynamics, and alignment objectives equally, in the absence of a prior favoring one term over the others.

The reconstruction loss is a measure of the difference between the original input and the final reconstructed output.
For each domain d, the reconstruction loss is defined as the mean squared error (MSE) between the input sequence $x_t^{(d)}$ and its reconstruction $\hat{x}_t^{(d)}$:

\begin{equation}
    \mathcal{L}_{\text{recon}}^{(d)} =
\mathbb{E}\!\left[\,\|x_t^{(d)} - \hat{x}_t^{(d)}\|_2^2\,\right],
\end{equation}

\noindent where $d$ is a domain variable,
$\hat{x}_t^{(d)}$ is a decoded output from a latent state.
The total reconstruction loss is computed as the average reconstruction error across both domains.

The dynamics loss models how the latent state evolves, given the current latent representation and control inputs.  
At each timestep, the dynamics head predicts the next latent state $\hat{z}_{t+1}$ based on the current latent $z_t$ and action vector $a_t$:
$\hat{z}_{t+1} = f_{\text{dyn}}(z_t, a_t)$,
where $z_t$ is the latent state produced by the encoder.

The dynamics loss is defined as the mean squared error between the predicted and true next latent states:
\begin{equation}
    \mathcal{L}_{\text{dyn}} =
    \mathbb{E}\!\left[\,\|z_{t+1} - \hat{z}_{t+1}\|_2^2\,\right].
\end{equation}

This encourages the latent representation to capture how its state evolves over time in response to control inputs, improving the model’s ability to capture motion dynamics across domains.  

To align the latent feature distributions of source and target platforms, we employ the Maximum Mean Discrepancy (MMD) loss, which measures the distance between the mean embeddings of two distributions in a reproducing kernel Hilbert space (RKHS) \cite{sriperumbudur2010hilbertspaceembeddingsmetrics}.

Given two sets of latent vectors X = $\{x_i\}_{i=1}^{m}$ and Y = $\{y_j\}_{j=1}^{n}$, sampled respectively from source and target distibutions, the squared MMD loss $L$ is defined as:

\begin{equation}
\mathcal{L}^2(P,Q) = \mathbb{E}_{P}[k(X,X)] - 2\,\mathbb{E}_{P,Q}[k(X,Y)] + \mathbb{E}_{Q}[k(Y,Y)]
\end{equation}

\noindent where $\mathbb{E}_{P}[k(X,X)]$ and $\mathbb{E}_{Q}[k(Y,Y)]$ denote expectations over all pairs of samples drawn from the same distribution, and $\mathbb{E}_{P,Q}[k(X,Y)]$ denotes the cross-domain expectation.
We use a Gaussian radial basis function (RBF) kernel:

\begin{equation}
    k(x,y) = \exp\!\left(-\frac{\|x - y\|^2}{2\sigma^2}\right),
\end{equation}
where $\sigma$ denotes the kernel bandwidth.

In practice, this loss minimizes the statistical distance between the U-CAT and Micro-CAT latent distributions. The concept of Maximum Mean Discrepancy and comparison with other kernel discrepancies is described in more detail in \cite{schrab2025practicalintroductionkerneldiscrepancies}. Training converges after 50 epochs.

\section{Data collection}

\begin{table}[t]
\centering
\caption{Robot platform specification}
\begin{tabular}{l|ll}
\textbf{Name} & \textbf{U-CAT} & \textbf{Micro-CAT} \\
\hline
Weight                & 22.0 kg      & 1.5 kg \\
Length                & 500 mm       & 150 mm \\
Hull diameter         & 250 mm       & 100 mm \\                       

Fin area              & 0.020 m$^2$   & 0.005 m$^2$ \\
Fin length            & 200 mm       & 110 mm \\
\hline
Camera                & Realsense D435i    & Raspberry Pi Cam V2 \\
IMU                   &Microstrain 3DM-CX5 & ICM-20608-G \\
Computer     &Jetson Orin Nano & Raspberry Pi Zero 2 W\\ & & + Arduino Pro Mini
\label{tab:robottab}
\end{tabular}
\end{table}

\begin{figure}
    \centering
    \includegraphics[width=0.9\linewidth]{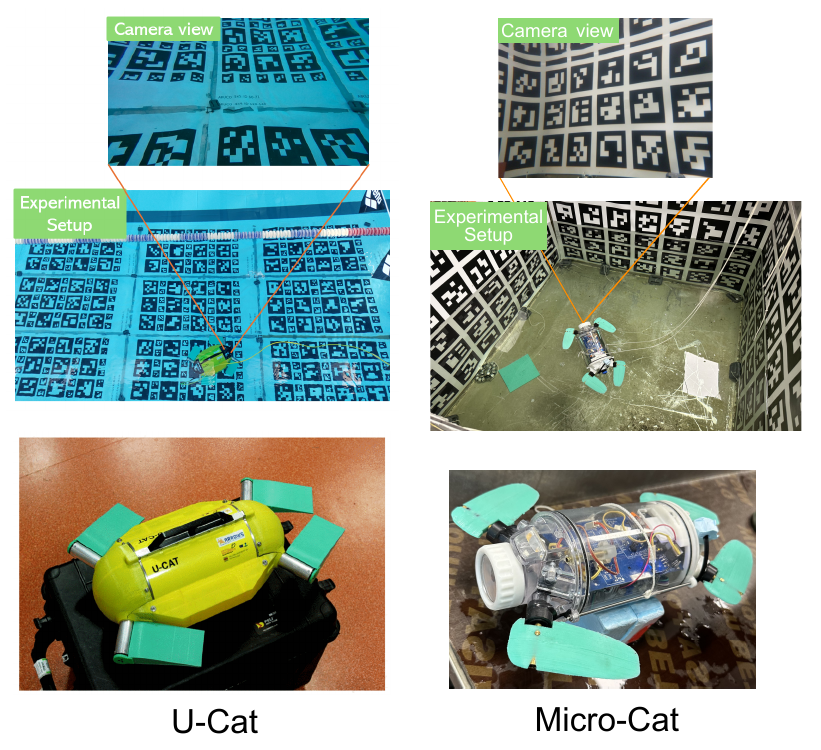}
    \caption{Experimental setup for both robots}
    \label{fig:exp_setup}
\end{figure}

\subsection{Robot specifications}\label{robots_specifications}


Two fin-actuated underwater robots, shown in Fig.~\ref{fig:exp_setup}, were used as platforms. We define morphological similarity as platforms sharing identical actuation kinematics and the same geometric design, while differing in absolute scale, the resulting mass and hydrodynamic properties. The two platforms satisfy this definition: both employ four independently controlled silicone fins with identical degrees of freedom, the same fin control equation (Eq. \ref{osc}), and share a common hull and fin layout with linear scale ratios of 3.33 in length, 2.5 in hull diameter, and 4.0 in fin area (Table~\ref{tab:robottab}). However the platforms differ in mass (ratio 14.67) and experience different hydrodynamic conditions, producing the domain gap. Thrust is generated by flapping motion regulated by amplitude $A^{amp}$, oscillation offset $\phi^{c}$, and frequency $f_{osc}$. The oscillatory movement for each fin at time $t$ is described by:

\begin{equation}
\label{osc}
\theta(t)=A^{amp} \sin \left(2\pi f_{osc} t\right)+\phi^{c}
\end{equation}

\noindent where $\theta(t)$ is set as the next target angle of the fin. These $(A^{amp}, \phi^{c})$ are included as input features for the modeling described in Section \ref{sec: method}. Fin oscillation amplitudes span $[0, 1.25]$ rad for U-CAT and $[0, 0.96]$ rad for Micro-CAT across the four flippers, while the lateral fins rotate through the full $[-\pi, \pi]$ rad range, and the vertical fins are constrained to $\pm 0.05$ rad. The trajectories include forward swimming, turning, and direction reversals. The dataset also includes Inertial Measurement Unit (IMU) measurements, with attitude estimated using an Extended Kalman Filter~\cite{valenti2015keeping}. Performance is reported in physical units (m/s) to preserve interpretability of platform-specific velocity scales, while normalized metrics could equally be used for cross-platform comparison.


\subsection{Experimental setup}{\label{Exp setup}}

Fig.~\ref{fig:exp_setup} shows the experimental setup. The robots were equipped with a camera to capture the robot's trajectory using ArUco markers \cite{garrido2014automatic} placed in the pool or the experimental water tank. These tags were positioned so that at least one was recognized within the robot camera's field of view, and pose estimation data was smoothed using a median filter. The velocity obtained by differentiating the position information obtained from the ArUco Markers was used as the ground truth for supervised learning and evaluation.

\begin{figure}[t]
    \centering
    \includegraphics[width=\linewidth]{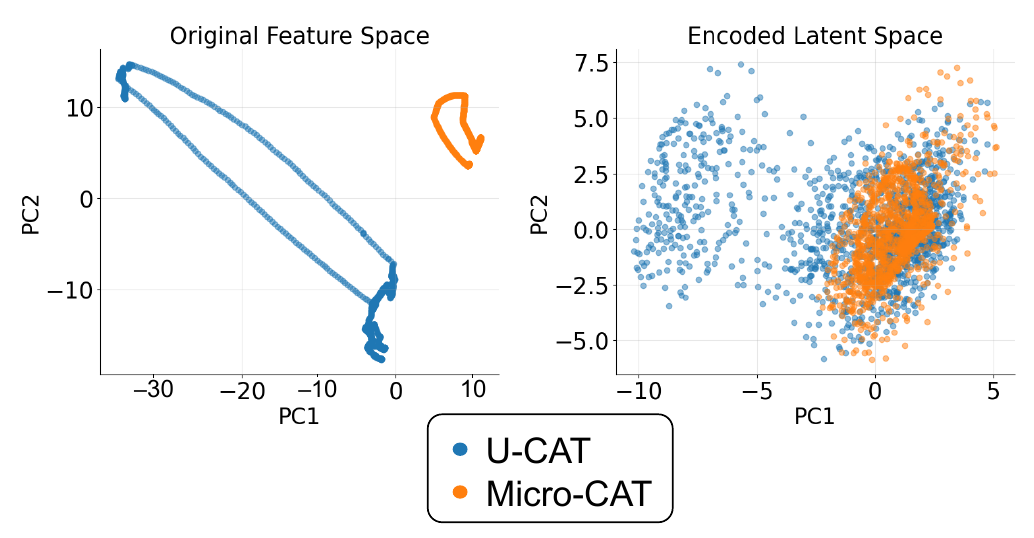}
    \caption{Data distributions before and after alignment. The increased overlap indicates successful reduction of the domain gap in latent space.}
    \label{fig:distribution_plots}
\end{figure}

\begin{figure*}[t]
    \centering
    \includegraphics[width=\linewidth]{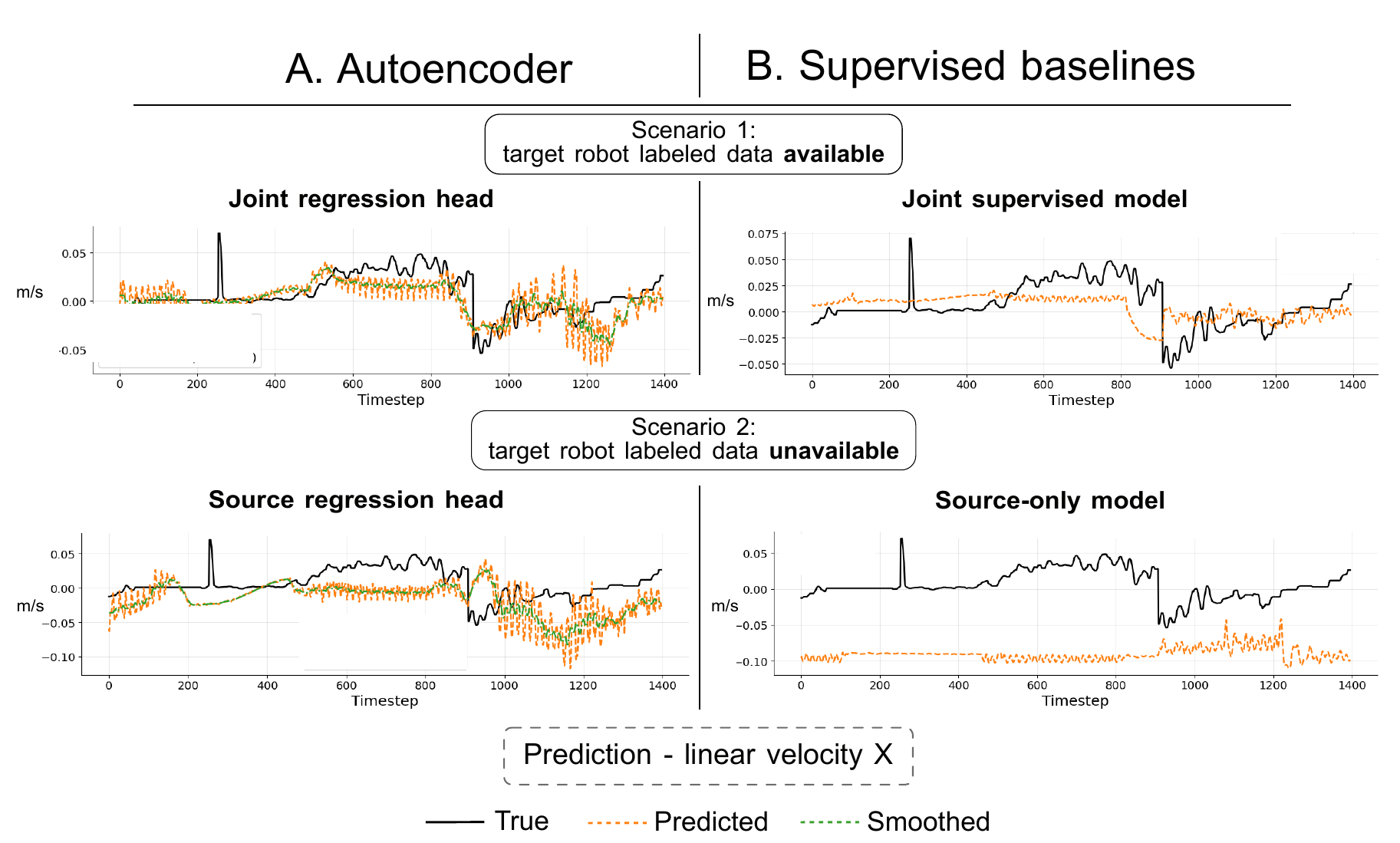}
    \caption{Comparison of predicted and ground-truth linear velocity in the X (forward) direction. The autoencoder follows the dominant trend in both scenarios, while the source-only baseline fails to generalize.}
    \label{fig:predictions_fig}
\end{figure*}

\begin{table*}[t]
\centering
\caption{Comparison of prediction performance across models for Micro-CAT test data. }
\label{tab:results_main}
\begin{tabular}{lcccccc}
\toprule
\textbf{Model} & \textbf{RMSE $V_x$} & \textbf{MAE $V_x$} & 
\textbf{RMSE $V_y$} & \textbf{MAE $V_y$} & 
\textbf{RMSE $\omega_z$} & \textbf{MAE $\omega_z$} \\
\midrule
Target-only (Micro-CAT) & 0.0328 & 0.0268 & 0.0270 & 0.0212 & 0.1013 & 0.0766 \\
Joint Supervised (U+M)  & 0.0237 & 0.0185 & 0.0232 & 0.0183 & 0.0861 & 0.0624 \\
Source-only (U-CAT)     & 0.0945 & 0.0888 & 0.0339 & 0.0288 & 0.1586 & 0.1156 \\
Autoencoder (Joint Head)      & 0.0362 & 0.0264 & 0.0288 & 0.0217 & 0.1151 & 0.0882 \\
Autoencoder (Source Head)     & 0.0596 & 0.0470 & 0.0678 & 0.0545 & 0.2309 & 0.1880 \\
\bottomrule
\end{tabular}
\end{table*}

\section{Results}

This section presents the results of dynamic prediction experiments comparing the proposed autoencoder with supervised baseline models. We first evaluate the autoencoder’s ability to align feature distributions across the two robot domains, followed by a comparison of prediction performance.

\subsection{Autoencoder Preliminary Results}

Fig.~\ref{fig:distribution_plots} shows the feature distributions of both robots before and after encoding. To assess latent alignment, we visualize the encoded features using Principal Component Analysis (PCA). PCA projects the high-dimensional latent representations onto the first two principal components, revealing the directions that explain most of the variation in the data. This visualization provides an interpretable view of how well the encoder aligns the two domains. As shown in Fig.~\ref{fig:distribution_plots}, the latent embeddings of the source (U-CAT) and target (Micro-CAT) become substantially more overlapping after training, indicating that the MMD loss effectively reduces the domain gap in the shared feature space.
For dynamic prediction evaluation, we consider the linear velocity components in $x$ and $y$ and the angular velocity around $z$. These components capture the primary motion of the robots, which includes forward–backward translation and yaw rotation. Prediction accuracy is measured using Root Mean Squared Error (RMSE) and Mean Absolute Error (MAE), averaged across three different test datasets.

\subsection{Evaluated models}{\label{sec:evaluated_models}}
The three supervised baselines defined in Section \ref{sec:baselines} serve as reference points.

To evaluate the learned latent representation, we use a linear probe approach \cite{zhang2016colorfulimagecolorization, chen2020simpleframeworkcontrastivelearning}: the frozen encoder produces features that are fed to a linear regression head, so predictive performance primarily reflects representation quality. Two variants are evaluated as shown in Fig. \ref{fig:overview}:  
an autoencoder with a joint head, trained on labels from both domains, and an autoencoder with a source head, trained on source labels only and evaluated zero-shot on the target.

\subsection{Prediction Comparison}
Fig.~\ref{fig:predictions_fig} illustrates how the autoencoder and baseline models predict the target robot’s linear velocity in the $x$-direction compared to ground truth. To smooth the autoencoder output, a moving average filter is applied to reduce high-frequency noise. The plots show that the autoencoder predictions follow the overall shape of the ground-truth velocity and capture sign changes, with some underestimation of short velocity peaks (near $t \approx 600-800$) and a small timing delay at sign changes (near $t \approx 1200$). The autoencoder preserves this behavior under both labeled and unlabeled target-domain conditions, whereas the joint-supervised model, and especially the source-only baseline, drift noticeably from the ground truth.
Table~\ref{tab:results_main} summarizes metrics results across all methods. The target-only model trained solely on labeled Micro-CAT data achieves strong performance, while the joint supervised model, trained on combined datasets, performs slightly better, confirming that exposure to multi-domain data improves generalization.  In contrast, the source-only model exhibits large errors, confirming the presence of a significant domain gap. The proposed autoencoder reduces this gap, improving zero-shot performance by $\sim$ 40\% in RMSE for Vx. The autoencoder with joint head achieves performance comparable to fully supervised baselines.

 These results demonstrate that while the joint supervised model achieves the lowest error, it requires fully labeled datasets from both platforms, which are costly and often impractical in underwater settings. In contrast, the proposed method achieves comparable performance without requiring labeled target data, making it more suitable for real-world deployment scenarios.


\section{Discussion}
The proposed approach does not outperform fully supervised joint training when labeled data from both domains is available. However, its primary advantage lies in label-efficient transfer, where target-domain annotations are limited or unavailable. Additionally, the current method assumes morphological similarity between platforms, and its effectiveness may decrease for significantly different robot designs. Extending the approach to more diverse robot morphologies may require more advanced domain adaptation techniques.
A systematic ablation of individual loss components - isolating the contribution of the MMD alignment and dynamics terms - would further characterize the framework and is left for future work. Similarly, a sensitivity analysis across $\lambda_{dyn}$ and $\lambda_{mmd}$, the RBF kernel bandwidth $\sigma$, and the latent dimension would help establish operational tolerances for transfer to new platforms.
Although the Maximum Mean Discrepancy (MMD) loss provided a stable and efficient domain alignment, future work may explore alternative alignment strategies, such as Kullback–Leibler (KL) divergence minimization \cite{cui2025generalizedkullbackleiblerdivergenceloss} or adversarial domain adaptation \cite{ganin2016domainadversarialtrainingneuralnetworks}. These methods could offer more flexibility in the implementation and tuning, but at the cost of higher training complexity. Extensions to recurrent or attention-based encoders, such as GRU, LSTM, Transformer, may further improve time-series-dependent modeling of dynamic features.

\section{Conclusion}
This study presented an efficient transfer learning framework for modeling the dynamics of morphologically similar underwater robots. By leveraging shared feature space of sensor data and control inputs, the proposed autoencoder-based model learns a common latent representation that aligns motion dynamics across platforms. Experimental results show that this approach achieves performance comparable to fully supervised baselines and enables reliable prediction for the smaller robot without labeled data, demonstrating effective zero-shot dynamics transfer.  

The findings suggest that morphological similarity helps transfer learned dynamics between robots, even when their size and hydrodynamic properties differ. Beyond underwater robots, the proposed framework can be applied to other soft or bio-inspired robots with comparable actuation principles, enabling scalable and label-efficient model transfer. Future work will extend this approach toward simulation-to-real adaptation and cross-platform control, including model-based and data-driven control strategies utilizing the shared latent dynamics representation.

\bibliographystyle{IEEEtran}
\bibliography{IEEEabrv,references}

\begin{thebibliography}{10}
\providecommand{\url}[1]{#1}
\csname url@samestyle\endcsname
\providecommand{\newblock}{\relax}
\providecommand{\bibinfo}[2]{#2}
\providecommand{\BIBentrySTDinterwordspacing}{\spaceskip=0pt\relax}
\providecommand{\BIBentryALTinterwordstretchfactor}{4}
\providecommand{\BIBentryALTinterwordspacing}{\spaceskip=\fontdimen2\font plus
\BIBentryALTinterwordstretchfactor\fontdimen3\font minus \fontdimen4\font\relax}
\providecommand{\BIBforeignlanguage}[2]{{%
\expandafter\ifx\csname l@#1\endcsname\relax
\typeout{** WARNING: IEEEtran.bst: No hyphenation pattern has been}%
\typeout{** loaded for the language `#1'. Using the pattern for}%
\typeout{** the default language instead.}%
\else
\language=\csname l@#1\endcsname
\fi
#2}}
\providecommand{\BIBdecl}{\relax}
\BIBdecl

\bibitem{biomimetics8030318}
G.~Li, G.~Liu, D.~Leng, X.~Fang, G.~Li, and W.~Wang, ``Underwater undulating propulsion biomimetic robots: A review,'' \emph{Biomimetics}, vol.~8, no.~3, 2023.

\bibitem{Zhang_2024_opt}
H.~Zhang, H.~Ma, Y.~Gao, Z.~Li, Y.~Chen, and Y.~Wang, ``Optimization of observational bionic underwater vehicle based on triz theory,'' \emph{Journal of Physics: Conference Series}, vol. 2795, no.~1, p. 012018, jul 2024.

\bibitem{zheng2022three}
X.~Zheng, M.~Xiong, R.~Tian, J.~Zheng, M.~Wang, and G.~Xie, ``Three-dimensional dynamic modeling and motion analysis of a fin-actuated robot,'' \emph{IEEE/ASME Transactions on Mechatronics}, vol.~27, no.~4, pp. 1990--1997, 2022.

\bibitem{wu2024review}
H.~Wu, Y.~Chen, Q.~Yang, B.~Yan, and X.~Yang, ``A review of underwater robot localization in confined spaces,'' \emph{Journal of Marine Science and Engineering}, vol.~12, no.~3, p. 428, 2024.

\bibitem{singh2025deepvldynamicsinertialmeasurementsbased}
M.~Singh and K.~Alexis, ``Deepvl: Dynamics and inertial measurements-based deep velocity learning for underwater odometry,'' 2025.

\bibitem{jaquier2024transferlearningroboticsupcoming}
N.~Jaquier, M.~C. Welle, A.~Gams, K.~Yao, B.~Fichera, A.~Billard, A.~Ude, T.~Asfour, and D.~Kragic, ``Transfer learning in robotics: An upcoming breakthrough? a review of promises and challenges,'' 2024.

\bibitem{muhammad2014non}
M.~R. Muhammad~Razif, N.~Elango, I.~N.~A. Mohd~Nordin, and A.~A. Mohd~Faudzi, ``Non-linear finite element analysis of biologically inspired robotic fin actuated by soft actuators,'' \emph{Applied Mechanics and Materials}, vol. 528, pp. 272--277, 2014.

\bibitem{singh2019dynamic}
N.~Singh, A.~Gupta, and S.~Mukherjee, ``A dynamic model for underwater robotic fish with a servo actuated pectoral fin,'' \emph{SN Applied Sciences}, vol.~1, no.~7, p. 659, 2019.

\bibitem{GEORGIADES200939}
C.~Georgiades, M.~Nahon, and M.~Buehler, ``Simulation of an underwater hexapod robot,'' \emph{Ocean Engineering}, vol.~36, no.~1, pp. 39--47, 2009, autonomous Underwater Vehicles.

\bibitem{lee2023data}
J.~Lee, K.~Viswanath, A.~Sharma, J.~Geder, M.~Pruessner, and B.~Zhou, ``Data-driven machine learning models for a multi-objective flapping fin unmanned underwater vehicle control system,'' in \emph{Proceedings of the AAAI Conference on Artificial Intelligence}, vol.~37, no.~13, 2023, pp. 15\,703--15\,709.

\bibitem{9652036}
G.~Li, T.~Stalin, V.~T. Truong, and P.~V.~y. Alvarado, ``{DNN}-based predictive model for a batoid-inspired soft robot,'' \emph{IEEE Robotics and Automation Letters}, vol.~7, no.~2, pp. 1024--1031, 2022.

\bibitem{hamamatsu2025underwater}
Y.~Hamamatsu, P.~Kupyn, R.~Gkliva, A.~Ristolainen, and M.~Kruusmaa, ``Underwater soft fin flapping motion with deep neural network based surrogate model,'' in \emph{2025 IEEE 8th International Conference on Soft Robotics (RoboSoft)}.\hskip 1em plus 0.5em minus 0.4em\relax IEEE, 2025, pp. 1--6.

\bibitem{Hamamatsu_2025}
Y.~Hamamatsu, W.~Remmas, J.~Rebane, M.~Kruusmaa, and A.~Ristolainen, ``Cross-platform learning-based fault tolerant surfacing controller for underwater robots,'' in \emph{2025 IEEE International Conference on Robotics and Automation (ICRA)}.\hskip 1em plus 0.5em minus 0.4em\relax IEEE, May 2025, p. 11263–11269.

\bibitem{UDP_transfer_learning2024}
Z.~Wu, Z.~Wu, X.~Chen, Y.~Lu, and J.~Yu, ``Self-supervised underwater image generation for underwater domain pre-training,'' \emph{IEEE Transactions on Instrumentation and Measurement}, vol.~73, pp. 1--14, 2024.

\bibitem{dastider2025crossembodimentroboticmanipulationsynthesis}
A.~Dastider, H.~Fang, and M.~Lin, ``Cross-embodiment robotic manipulation synthesis via guided demonstrations through cyclevae and human behavior transformer,'' 2025.

\bibitem{sriperumbudur2010hilbertspaceembeddingsmetrics}
B.~K. Sriperumbudur, A.~Gretton, K.~Fukumizu, B.~Schölkopf, and G.~R.~G. Lanckriet, ``Hilbert space embeddings and metrics on probability measures,'' 2010.

\bibitem{schrab2025practicalintroductionkerneldiscrepancies}
A.~Schrab, ``A practical introduction to kernel discrepancies: Mmd, hsic \& ksd,'' 2025.

\bibitem{valenti2015keeping}
R.~G. Valenti, I.~Dryanovski, and J.~Xiao, ``Keeping a good attitude: A quaternion-based orientation filter for imus and margs,'' \emph{Sensors}, vol.~15, no.~8, pp. 19\,302--19\,330, 2015.

\bibitem{garrido2014automatic}
S.~Garrido-Jurado, R.~Mu{\~n}oz-Salinas, F.~J. Madrid-Cuevas, and M.~J. Mar{\'\i}n-Jim{\'e}nez, ``Automatic generation and detection of highly reliable fiducial markers under occlusion,'' \emph{Pattern Recognition}, vol.~47, no.~6, pp. 2280--2292, 2014.

\bibitem{zhang2016colorfulimagecolorization}
R.~Zhang, P.~Isola, and A.~A. Efros, ``Colorful image colorization,'' 2016.

\bibitem{chen2020simpleframeworkcontrastivelearning}
T.~Chen, S.~Kornblith, M.~Norouzi, and G.~Hinton, ``A simple framework for contrastive learning of visual representations,'' 2020.

\bibitem{cui2025generalizedkullbackleiblerdivergenceloss}
J.~Cui, B.~Zhu, Q.~Xu, Z.~Tian, X.~Qi, B.~Yu, H.~Zhang, and R.~Hong, ``Generalized kullback-leibler divergence loss,'' 2025.

\bibitem{ganin2016domainadversarialtrainingneuralnetworks}
Y.~Ganin, E.~Ustinova, H.~Ajakan, P.~Germain, H.~Larochelle, F.~Laviolette, M.~Marchand, and V.~Lempitsky, ``Domain-adversarial training of neural networks,'' 2016.

\end{thebibliography}

\end{document}